# MuSTEL: A Large-Scale Multilingual Synthetic Text-Line Image Dataset for OCR

Majid Adibian[1,2,†], Hossein Saberi[2,†], Hossein Zeinali[1,*]

## Abstract

Optical character recognition (OCR) systems depend strongly on the scale and diversity of their training data. One approach to document OCR first analyzes the page layout and extracts text-line images, which are then transcribed by a recognition model. Multilingual line-level recognition involving Persian, Arabic, and English presents additional challenges because different writing directions, fonts, and styles may coexist within a single line. This paper introduces MuSTEL, a large-scale synthetic multilingual OCR dataset containing more than 14.5 million annotated text-line images, organized into a training partition and predefined small and large test sets. Samples are generated from natural-language corpora, including multilingual content already present in the source texts. The generation framework incorporates capability-aware font selection, within-line font and style changes, mixed-direction text rendering, highlighting, and underlining. A validated repository of 2,009 font files from 711 families supports broad typographic coverage, while varied backgrounds, blur, geometric distortion, skew, stroke width, and image save quality provide diverse document-like appearances. Statistical analysis characterizes the dataset's language composition, transcription lengths, lexical and character-sequence diversity, image dimensions, typography, and rendering conditions. MuSTEL provides a resource for training and evaluating multilingual line-level OCR systems and investigating recognition robustness across languages, writing directions, and visual conditions.

## 1. Introduction

Optical character recognition (OCR) converts text in images into machine-readable form and supports applications such as digital archiving [1], document management [2], information extraction [3], historical-document preservation [4], and intelligent search [5]. Although deep learning has substantially improved OCR performance, model effectiveness and generalization still depend strongly on the scale, quality, and diversity of training data [6].

* Corresponding author.
*E-mail address:* hzeinali@aut.ac.ir (Hossein Zeinali).
† Contributing authors.
*E-mail address:* majid77@aut.ac.ir (Majid Adibian).
*E-mail address:* hossein.saberi1998@gmail.com (Hossein Saberi).
[1] Department of Computer Engineering, Amirkabir University of Technology, Tehran, Iran
[2] Sharif DeepMine (Zharf Kavan Sharif) Co., Tehran, Iran

Some OCR systems recognize text directly from complete documents or images [7], while others first detect or segment text lines and recognize them independently. The latter approach requires line-level training data that capture variation in language, writing direction, typography, and image appearance [8]. These requirements motivate MuSTEL, a large-scale synthetic multilingual dataset for Persian, Arabic, and English line-level OCR [9].

OCR tasks differ in both visual input and recognition granularity. Scene-text recognition addresses text in natural environments, where perspective, illumination, complex backgrounds, and irregular placement affect readability [10]. Document OCR processes scanned, photographed, or digitally generated documents and may operate at the page, region, word, or text-line level [11].

Line-level recognition predicts the character sequence of a single text-line image. By separating transcription from page-layout analysis, it supports pipelines that detect or segment lines before recognizing their content. This setting is particularly relevant to books, reports, archives, and other text-dense documents [8].

Large annotated datasets help recognition models learn variations in character appearance, typography, textual context, and image quality [12]. Synthetic generation makes it possible to produce accurately labeled text-line images at a scale that would be costly and time-consuming to achieve through manual annotation [13]. For multilingual document OCR, these images should also represent diverse scripts, fonts, styles, and visual conditions [14].

Multilingual recognition becomes more challenging when different writing systems coexist within a line [9]. Persian and Arabic use right-to-left cursive scripts with contextual character forms and joining behavior, whereas English is written from left to right [15]. Although Persian and Arabic share many characters, they retain distinct letters and orthographic conventions [16]. Embedded English content therefore requires recognition systems to handle changes in writing direction while preserving the correct transcription order [14].

The challenge extends beyond language switching. Document lines may contain changes in font family, style, size, and emphasis, alongside numerals, punctuation, and technical expressions. Background variation, blur, skew, distortion, and compression introduce further visual complexity. Training resources should therefore represent linguistic, typographic, and visual variability together [17].

Existing datasets address these requirements only partially. Many widely used resources focus on English or other Latin-based languages, while resources jointly covering Persian, Arabic [10], and English at the line level remain limited [18]. Restricted font collections, predominantly single-font rendering, and limited formatting or appearance variation further motivate a dataset combining multilingual content with broad typographic and visual diversity [19].

To address these limitations, MuSTEL provides synthetic text-line images for Persian, Arabic, English, and mixed-language recognition. Its main contributions are:

- **Large-scale multilingual line-level OCR dataset:** MuSTEL contains more than 14.5 million annotated images, including mixed-language content already present in the source corpora. Predefined training, small-test, and large-test partitions support evaluation at different computational scales.
- **Validated multilingual typography and capability-aware rendering:** A manually validated repository of 2,009 font files from 711 families supports font selection according to the required scripts, numerals, punctuation, hamzah forms, and diacritics, together with within-line font and style changes.
- **Linguistic and stylistic diversity:** The generation framework extends TRDG to render monolingual and mixed-language lines with varied typography, highlighting, and underlining. Backgrounds, blur, distortion, skew, stroke width, and image save quality provide additional document-like variation.

The remainder of this paper is organized as follows: Section 2 reviews related work, Section 3 describes dataset construction, Section 4 presents statistical analysis, Section 5 discusses advantages, applications, limitations, and future directions, and Section 6 concludes the paper.

## 2. Related Work

The progress of deep learning-based OCR has been closely linked to the availability of large annotated datasets [6]. Existing resources differ substantially in language coverage [10], recognition granularity [11], typographic diversity [20], and visual realism [21], while synthetic generation has become an important strategy for producing large-scale labeled data [17]. This section reviews representative OCR datasets and generation approaches relevant to multilingual OCR and summarizes the remaining limitations.

OCR datasets differ along two dimensions: the source of the images and the unit of text they annotate [22]. Scene-text datasets contain text photographed in natural environments, where perspective, lighting, and surrounding objects affect its appearance [10]. Document-oriented datasets contain scanned, photographed, or digitally generated pages and text crops [23]. Their annotations may correspond to characters, words, lines, regions, or entire pages; datasets at these levels support different recognition tasks and are not directly interchangeable [11].

This distinction matters when assessing resources for line-level OCR. A page-level dataset requires line segmentation and corresponding transcriptions before it can train a line recognizer, while isolated word images do not capture the full composition of a document line [24, 25]. Line-level datasets pair each cropped line with its transcription, allowing evaluation of complete sequences, including punctuation, spaces, and transitions between scripts or fonts [8, 9].

English OCR research has benefited substantially from large synthetic datasets. MJSynth introduced millions of rendered English word images generated from a large vocabulary and diverse fonts [26]. SynthText extended synthetic generation by embedding text into natural scenes, accounting for scene geometry, perspective, and illumination [27]. These resources demonstrated the effectiveness of large-scale synthetic data for training text-recognition models. However, they primarily target English scene-text and word-level recognition and provide limited representation of multilingual document lines, right-to-left scripts, and within-line language or typography changes [10].

Dedicated datasets have supported OCR research for both Arabic and Persian, but their recognition units, scale, and typographic coverage differ. For Arabic, BCE-Arabic-v1 provides 1,833 scanned pages from 180 books with varied layouts and page content, supporting document layout analysis and OCR [28]. APTI instead provides synthetic Arabic word images generated using ten fonts, ten sizes, and four styles [29]. For Persian, Shotor contains 120,000 synthetic word images, whereas IDPL-PFOD provides 30,138 synthetic text-line images rendered using eleven fonts, with variations in style, size, background, blur, and distortion [30].

More recent resources such as SARD moves toward large-scale document-oriented synthesis and realistic page layouts [31]. However, most existing Arabic and Persian datasets remain language-specific, and support for multilingual composition, extensive font diversity, and Persian-Arabic-English line-level recognition remains limited [18].

Multilingual datasets have broadened language coverage in text detection and recognition benchmarks. ICDAR2017 MLT contains 18,000 natural-scene images across nine languages for text detection and script identification [32]. ICDAR2019 MLT extends this benchmark to 20,000 images across ten languages and adds end-to-end recognition [33]. Both include Arabic and English but exclude Persian from their specified language inventories. $M^6$Doc provides Chinese and English document pages primarily for layout analysis; this language combination does not introduce transitions between right-to-left and left-to-right scripts [34].

Multilingual coverage therefore does not necessarily entail changes in writing direction within a recognition sequence [34]. Although MLT includes Arabic and mixed-script instances, it does not specifically target bidirectional Persian-Arabic-English document lines [33]. MuSTEL addresses this setting by combining the three languages within individual lines, including direction changes between Persian or Arabic and embedded English, together with controlled font-family and style variation.

Synthetic data generation has become an important strategy for OCR because collecting and manually annotating large document datasets is costly and time-consuming, particularly for low-resource languages and multilingual settings. Synthetic approaches can produce large quantities of accurately labeled samples while providing direct control over textual content, typography, backgrounds, and image appearance [6, 35].

**Table 1.** Comparative analysis of representative OCR datasets with respect to language coverage, dataset characteristics, and multilingual capabilities.

| Dataset | Languages | Images | Synth. | Mixed Languages / Multi-Font | Font diversity / Augmentations | Public / Year | Notes |
|---|---|---|---|---|---|---|---|
| APTI | Arabic | ~45M (words) | Yes | No / No | Fonts: 10 Fonts Aug.: None (clean synthetic) | Yes / 2009 | Synthetic words; ~250 M characters total. |
| MJSynth (Synth90k) | English | 9M | Yes | No / No | Fonts: High Aug.: Limited (plain text) | Yes / 2014 | 90K-word English vocabulary; one word per image, with multiple rendered samples per word. |
| SynthText | English | ~800K | Yes | No / No | Fonts: High Aug.: Scene-text backgrounds | Yes / 2016 | Synthetic text rendered on natural image backgrounds. |
| BCE-Arabic-v1 | Arabic | 1833 (pages) | No | No / No | Fonts: Medium (~10 fonts) Aug.: None | Yes / 2016 | Scanned Arabic book pages (scanned page images). |
| AHCD (El-Sawy et al.) | Arabic | ~17K | No | No / No | Fonts: Unknown Aug.: None | Yes / 2017 | Arabic handwritten character images. |
| Hesham et al. (Univ. study) | Arabic | 85 (pages) | No | No / No | Fonts: Unknown Aug.: None | No / 2017 | Arabic document pages (unpublished data). |
| RDI 109-docs (RDI Egypt) | Arabic | 109 (pages) | No | No / No | Fonts: Unknown Aug.: None | No / 2018 | Proprietary book/magazine pages (RDI CleverPage). |
| Arabic-Docs (Kaggle) | Arabic | 10K (pages) | No | No (mix print/hand) / No | Fonts: Many fonts (various) Aug.: None (real scans) | Yes / 2019 | 12 classes (invoices, books, etc.), mixed content. |
| Shotor | Persian | 120K | Yes | No / No | Fonts: Medium (multiple fonts) Aug.: Limited | Yes / 2020 | Synthetic 50×100 word images from Wikipedia text. |
| IDPL-PFOD | Persian | 30K | Yes | No / No | Fonts: Medium (common Farsi fonts) Aug.: Yes (blur, distortions) | Yes / 2021 | 700×50px text-line images (10% with background noise). |
| SARD | Arabic | ~3M pages | Yes | No / No | Fonts: 6 Fonts Aug.: Yes (clean rendering; book-style layouts) | Yes / 2026 | Synthetic book-style pages containing 794.6M words; page-level image–transcription pairs |
| **MuSTEL (This Study)** | Persian, Arabic, English | 14.5M+ | Yes | Yes / Yes | Fonts: 2009 fonts files; 711 families Aug.: Yes (document-like) | Yes / 2026 | Monolingual and mixed-language text lines with within-line font and style changes. |

Early methods mainly rendered isolated words or text lines using selected fonts and simple backgrounds, while later approaches introduced geometric and appearance transformations to improve realism [36]. General-purpose frameworks such as Text Recognition Data Generator (TRDG) provide configurable fonts, colors, backgrounds, spacing, and distortions. However, its standard generation workflow uses a single font per image and does not provide dedicated mechanisms for underlining, highlighting, or controlled language and font-style switching within a line [37]. MuSTEL extends TRDG by implementing these features, enabling mixed-language lines with changes in font family and style, together with underlining and highlighting.

Despite these advances, additional customization is often required for multilingual Persian or Arabic-script OCR, particularly for right-to-left rendering [38], script-aware font selection [39], mixed-language composition, and multiple-font text lines [40].

Table 1 lists representative OCR datasets and compares their language coverage, scale, and rendering characteristics. None of the listed datasets jointly covers Persian, Arabic, and English. MuSTEL addresses this gap with more than 14.5 million text-line images, a validated repository of 711 font families, and varied document-like rendering conditions. Its distinction lies in combining this scale and three-language coverage with mixed-language composition and controlled changes in language, font family, and style within individual lines.

## 3. Dataset Construction

MuSTEL was constructed as a large-scale synthetic line-level OCR dataset for Persian, Arabic, English, and mixed-language text. The construction framework combines multilingual natural-language corpora, a capability-validated font repository, language-aware text composition, script-aware rendering, and diverse document-like visual conditions. It extends TRDG to support the multilingual and typographic requirements of the dataset.

The following subsections describe the data-generation pipeline, font validation, and generation of text-line images with diverse linguistic, typographic, and visual properties.

### 3.1. Overview of the Dataset Generation Pipeline

The MuSTEL generation pipeline transforms source text and validated fonts into annotated text-line images. As illustrated in Figure 1, it involves text sampling, font and style selection, text rendering, and appearance variation, including image save quality. Source text is preserved without normalization, and mixed-language lines originate directly from multilingual content already present in the source corpora.

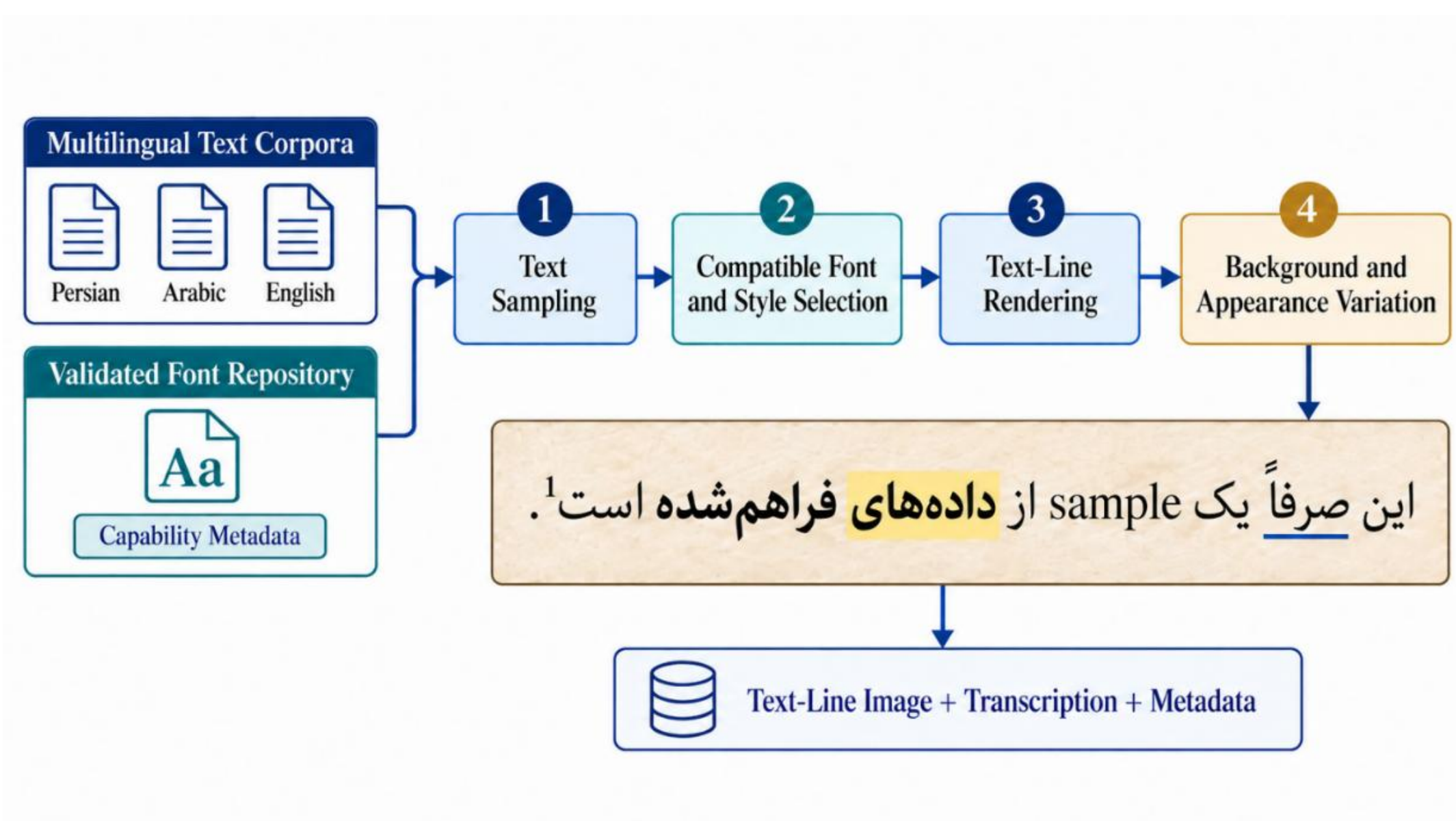

**Figure 1.** Overview of the MuSTEL dataset generation pipeline.

Text is first sampled from the prepared Persian, Arabic, or English corpora. Each sample may contain one or more languages, reflecting the monolingual or multilingual content already present in the source text and preserving its original sequence.

The pipeline then selects compatible fonts and styles from the validated repository and renders the text while preserving script-specific shaping and bidirectional behavior. Both single and multiple-font configurations are supported.

Background and image-appearance parameters are subsequently sampled to introduce document-like variation. The generated line image, transcription, and associated metadata then undergo quality-control checks before the sample is accepted.

## 3.2. Font Collection and Validation

Typographic diversity is particularly important for Persian and Arabic OCR because typefaces differ in character shapes, joining behavior, ligatures, stroke width, and spacing. MuSTEL therefore uses a curated multilingual font repository covering Persian, Arabic, and English.

The fonts were collected from three principal sources: Google Fonts, ArabicFonts.net, and FontYab.com. Google Fonts provided a broad collection of Latin and multilingual families, ArabicFonts.net expanded Arabic typography, and FontYab contributed additional Persian-oriented typefaces. The downloaded resources were parsed to extract metadata including family name, style, file format, Unicode coverage, and available font faces. Font families, files, and faces were treated separately because a single family may contain multiple variants such as Regular, Bold, Italic, Light, or Condensed.

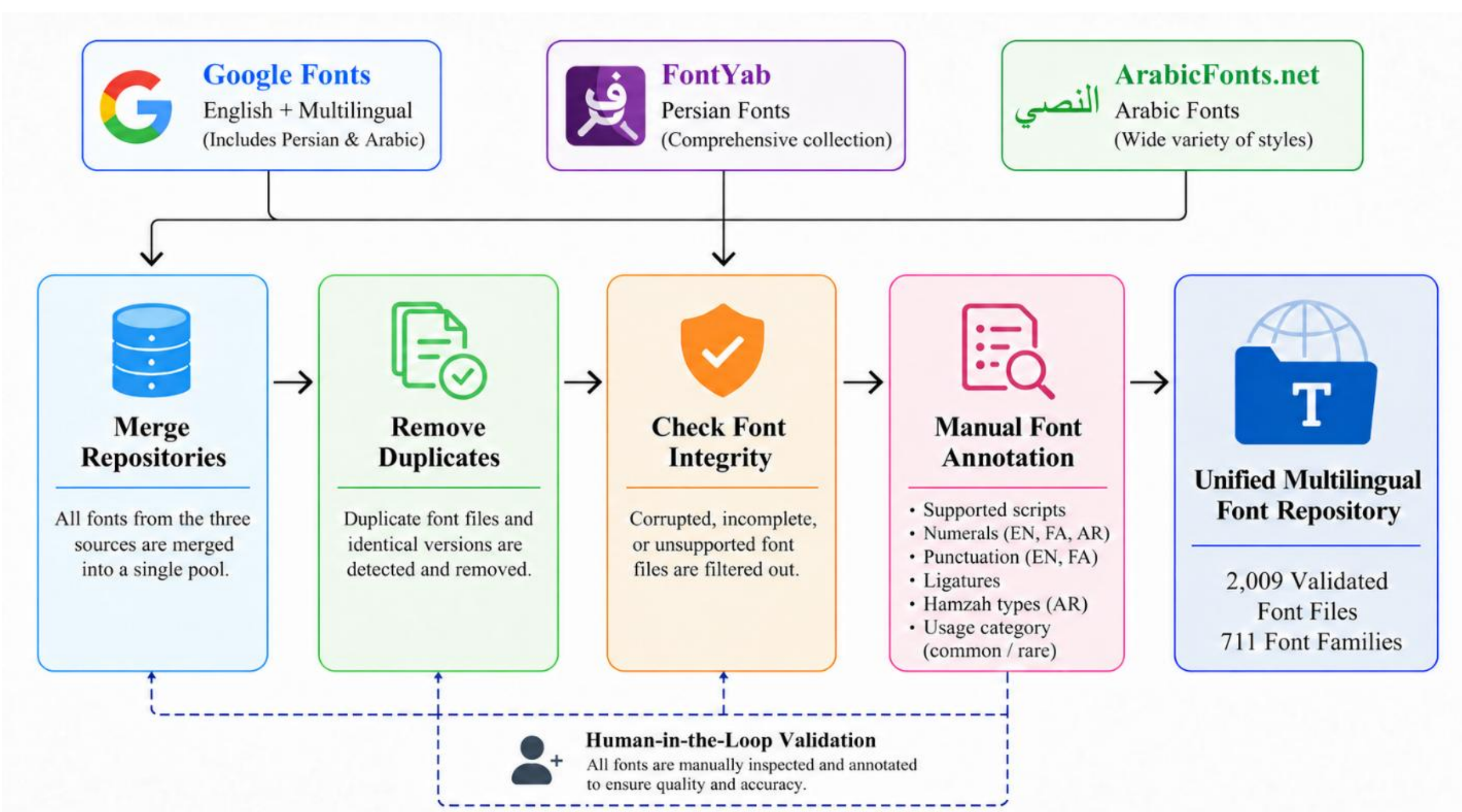

**Figure 2.** Workflow of the multilingual font collection and repository construction process.

A multistage filtering procedure was applied to reduce redundancy and remove unusable resources. Byte-identical files were detected using cryptographic hashes, while normalized family and style names, character maps, and font-table information were used to identify likely duplicates. Fonts with corrupted structures, unsupported formats, or persistent loading and rendering failures were excluded. Partial character coverage did not automatically result in removal, since a font could still be used for compatible text segments. Common and less-common typefaces were also distinguished so that frequently encountered document fonts could be sampled more often while retaining less-common designs for additional diversity. The overall font collection, filtering, validation, and repository construction workflow is summarized in Figure 2.

Because Unicode character maps do not guarantee correct visual rendering, the remaining fonts underwent both automated and manual validation. Automated checks verified structural integrity and candidate support for predefined Persian, Arabic, and English character groups. A diagnostic image was then generated for each font face and style using representative English letters, Persian and Arabic characters, numeral systems, punctuation, hamzah forms, diacritics, and joining patterns.

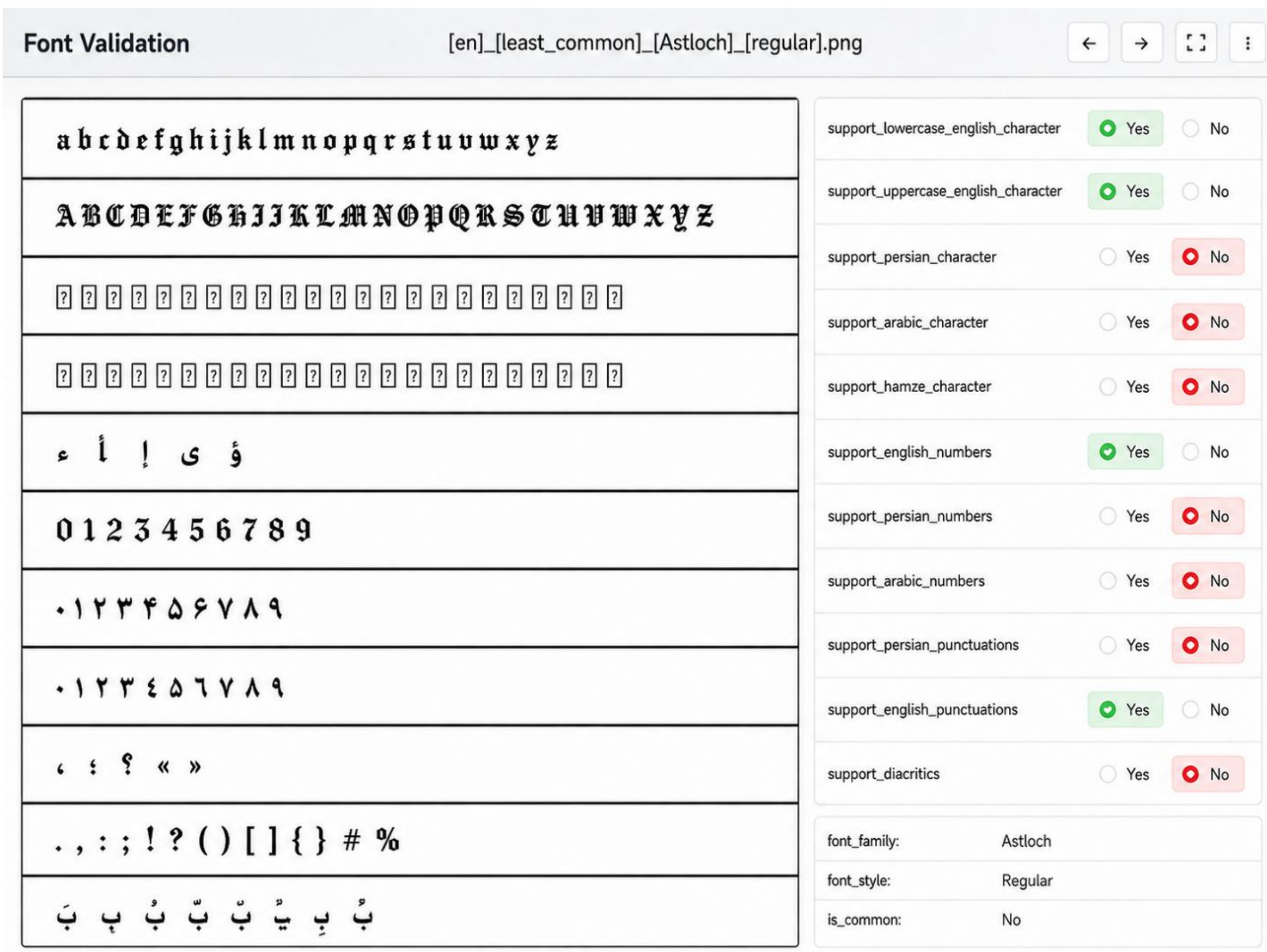

**Figure 3.** Examples of manual font validation and annotation in the MuSTEL font repository.

The diagnostic images were reviewed in Label Studio by a team of ten annotators. For each font face, reviewers recorded the family and style and assigned capability-specific labels indicating whether the evaluated character groups were rendered correctly. A capability was accepted only when the relevant characters appeared without missing-glyph symbols, corrupted outlines, unintended substitutions, severe clipping, or visually incorrect forms. For Persian and Arabic, validation additionally considered contextual shaping, joining behavior, ligatures, hamzah forms, and combinations containing diacritics. Representative examples of the manual font-validation and annotation interface are shown in Figure 3.

Capabilities were recorded independently rather than assigning a single language-level pass or fail. Fonts with partial support could therefore remain in the repository and be selected only for compatible text segments. Uncertain capabilities were conservatively treated as unsupported. The resulting font-capability metadata is queried during generation so that each text run is assigned only fonts validated for its required scripts, numerals, punctuation, and other character groups.

## 3.3. Text Corpus Collection and Preprocessing

MuSTEL uses natural-language corpora rather than random character strings or isolated word lists to preserve realistic vocabulary, punctuation, spacing, and contextual character sequences in the generated text lines.

Separate corpora were assembled for Persian, Arabic, and English using two principal source types: Wikipedia articles and digitized books. Wikipedia provided broad lexical and topical coverage across scientific, historical, cultural, geographical, and general-domain subjects, while the book collections contributed longer and more structured passages from literature, education, science, history, and related document domains. Combining these sources increased vocabulary diversity and reduced dependence on the linguistic characteristics of a single corpus.

The source documents were processed separately for each language. Markup, references, navigation elements, document-structure artifacts, and other non-content material were removed where applicable. Exact and normalized duplicate lines were also removed to reduce unnecessary repetition.

Zero-width non-joiners were preserved in Persian text to retain the original word structure and joining behavior. Tatweel characters were additionally introduced into selected Persian and Arabic samples to create controlled variation in character elongation during rendering.

Finally, long texts were divided into smaller segments of randomly selected lengths for line-level image generation. Each segment could contain part of a sentence rather than a complete sentence. The Persian, Arabic, and English corpora were retained separately to preserve their source-language identities.

### 3.4. Synthetic Line-Image Generation

The MuSTEL image-generation framework was developed by extending the open-source TRDG with multilingual and typography-aware rendering components required for Persian, Arabic, and English line-level OCR. In particular, the extended framework supports script-specific shaping, bidirectional composition, language-aware font selection, controlled within-line changes in font family and style, and additional formatting features such as underlining and highlighting.

For each sample, the text is first analyzed to determine its language and script composition, writing directions, grapheme structure, and required font capabilities. Depending on the selected generation mode, the line may contain Persian, Arabic, English, or a combination of two or three languages. Mixed-language lines originate directly from multilingual content already present in the source corpora, preserving the original text sequences and their logical order. Persian and Arabic sequences are treated as right-to-left content, whereas English sequences retain their left-to-right direction. Bidirectional layout is resolved during rendering so that embedded English terms, numerals, and other mixed-script elements appear in the appropriate visual positions while the stored ground-truth transcription remains in logical order.

Font selection is constrained by the capability annotations described in *Section 3.2. Font Collection and Validation*. Before a rendering run is assigned a font, its required characters are mapped to the corresponding capability groups, including language-specific letters, numerals, punctuation, hamzah forms, and diacritics. Candidate fonts are restricted to those validated for the complete requirements of the run. A final rendering check is performed before acceptance, and an alternative compatible font is selected if the candidate fails.

MuSTEL supports both single-font and multiple-font line generation. For multiple-font samples, the transcription is divided into contiguous rendering runs at valid grapheme boundaries so that base characters are not separated from dependent marks or diacritics. Each generated sample may use one or more compatible font families and styles across different text segments. Font family and style are selected separately, allowing a style change within the same family or a font-family change while retaining the same style.

Persian and Arabic runs undergo contextual shaping before rendering so that the appropriate connected forms and supported ligatures are produced. Because fonts may differ in ascent, descent, glyph height, and internal metrics, independently rendered runs are aligned to a shared baseline before being composited into the final line image. Canvas dimensions are calculated from the combined rendered extents with sufficient margins to reduce clipping of dots, diacritics, descenders, and other typographic elements. Samples exhibiting missing glyphs, malformed shaping, severe clipping, or invalid composition are rejected.

The complete generation procedure is summarized in Algorithm 1, which integrates language-composition sampling, capability-aware font assignment, grapheme-safe rendering, contextual shaping, bidirectional layout, appearance parameter sampling, image validation, and metadata storage.

**Algorithm 1:** MuSTEL Multilingual Synthetic Text-Line Generation.

**Input:** Text corpora $\mathcal{C}$, validated font repository $\mathcal{F}$, font capabilities $A$, generation configuration $P$, target size $N$.

**Output:** Synthetic dataset $\mathcal{D}$.

$\mathcal{D} \leftarrow \emptyset$
**while** $|\mathcal{D}| < N$ **do:**
  Sample a text $T$ from $\mathcal{C}$.
  Randomly split $T$ into $K \in \{1,2\}$ nonempty grapheme-aligned runs $\{r_j\}_{j=1}^{K}$.
  Identify compatible fonts for each run: $C_j = \{f \in F \mid R(r_j) \subseteq A(f)\}$.
  **if** any $C_j = \emptyset$ **then continue**
  Randomly select $f_j \in C_j$ for each run $r_j$.
  Sample rendering and appearance parameters $\theta \sim P$.
  $I_0 \leftarrow \mathrm{Render}_{BiDi}(\{(r_j, f_j)\}_{j=1}^{K}, \theta)$
  Compose background and text formatting: $I_1 \leftarrow \mathrm{Compose}(I_0, \theta)$
  Apply image transformations: $I \leftarrow \mathrm{Augment}(I_1, \theta)$
  **if** $\mathrm{Valid}(I, T)$ **then:**
    $\mathcal{D} \leftarrow \mathcal{D} \cup \{(I, T, M)\}$
  **end if**
**end while**
**return** $\mathcal{D}$

**Notation:** $R(r_j)$ denotes the character capabilities required by run $r_j$; $A(f)$ denotes the validated capabilities of font $f$. $C_j$ contains the compatible fonts for run $r_j$. $K$ denotes the number of rendering runs; the same font may be selected for multiple runs. The generation metadata $M$ records the selected fonts, text labels, and rendering parameters.

## 3.5. Diversity in Data Generation and Quality Control

Text lines extracted from practical documents can exhibit substantial visual variation because of differences in typography, background appearance, document production, scanning, and image storage. To represent such conditions, MuSTEL varies rendering parameters during image generation rather than producing all samples with a uniform visual configuration. These parameters include font size, text appearance, margins, background type, blur, skew, geometric distortion, stroke width, image save quality, highlighting, and underlining.

Background variation contributes to visual diversity, and MuSTEL uses several background types, including a collection of pre-stored images. Figure 4 illustrates background variation alongside changes in typography, text geometry, and formatting.

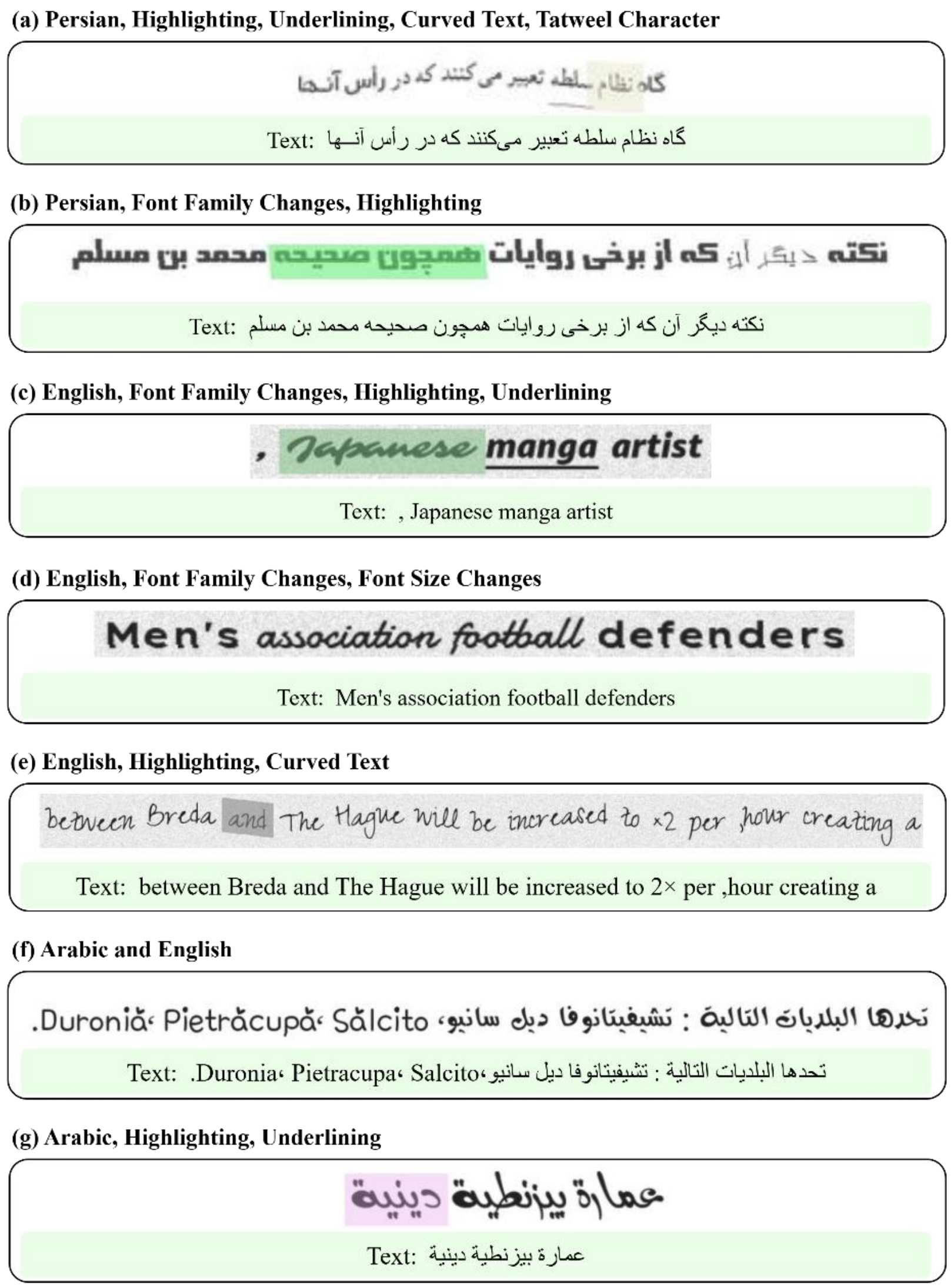

**Figure 4.** Examples of multilingual text-line rendering, font variation, and text formatting in MuSTEL.

Additional parameters control blur, skew, geometric distortion, stroke width, image save quality, highlighting, and underlining. These characteristics are sampled according to the generation configuration and may occur jointly within the same text-line image.

Selected samples also include manually inserted superscript and specialized content, such as national identification numbers, telephone numbers, and table-of-contents entries. These cases are identified by dedicated flags in the metadata, while superscript content additionally uses a special transcription marker.

Rendering variations are applied while each text-line image is generated. Each sample is created with its own combination of text, fonts, background, geometric effects, and image-quality settings, rather than by modifying an existing image to produce additional copies. These settings are recorded in the sample metadata for subsequent analysis.

MuSTEL uses automatically generated annotations because both the textual content and rendering configuration are known during image synthesis. Each accepted text-line image is associated with its ground-truth transcription, stored as UTF-8 text in logical character order. For mixed-direction samples, the annotation preserves the underlying transcription sequence rather than the visually arranged glyph order. Contextual Persian and Arabic glyph forms and ligatures are represented using their corresponding canonical Unicode sequences rather than presentation-form code points.

Each sample is assigned a unique identifier linking the image to its transcription and associated metadata. Recorded metadata include language composition, font family and style, and generation parameters such as background type, spacing, margins, blur, skew, distortion, image quality, highlighting, and underlining.

The rendering configuration is checked before image generation, and an image is produced if the process completes without an error. Images are saved at different quality settings to introduce variation in their appearance. Each saved image is linked to its transcription and generation metadata for use in OCR training and dataset analysis.

### 3.6. Dataset Organization and Splits

MuSTEL is organized into a large training set and two predefined evaluation sets. The training set contains 14,377,989 text-line images and serves as the principal resource for model development. A small test set of 6,000 images, with 2,000 samples from each primary language, provides a resource-efficient option for rapid evaluation. A large test set of 143,999 images supports more comprehensive assessment and is likewise approximately balanced across English, Persian, and Arabic.

Two predefined test sets are provided to accommodate different evaluation budgets. The small test set contains 6,000 images, with 2,000 samples from each primary language, and enables rapid evaluation when computing resources are limited or OCR services charge per image. The large test set contains 143,999 images, approximately balanced across the three languages, and supports a more comprehensive assessment using substantially more samples. Researchers can therefore select the evaluation scale according to their computational and financial resources. Partition labels are retained in the dataset metadata to enable consistent comparisons across studies, and detailed language counts are reported in *Section 4.1*.

## 4. Dataset Statistics and Analysis

Dataset scale alone does not determine the usefulness of an OCR resource; linguistic, typographic, lexical, and visual diversity also help represent the conditions encountered in practical recognition tasks. MuSTEL was therefore analyzed across several dimensions, including language composition, transcription length, image dimensions, lexical and character-sequence coverage, font and style usage, and rendering characteristics.

## 4.1. Overall Dataset Statistics

MuSTEL contains 14,527,988 synthetic line-level text images generated from Persian, Arabic, and English textual resources. The dataset includes 4,853,510 English samples, 4,825,564 Persian samples, and 4,848,914 Arabic samples, corresponding to 33.41%, 33.22%, and 33.38% of the complete collection, respectively. The close proportions of the three primary languages provide a broadly balanced multilingual dataset.

The released data consist of a large training set and two evaluation sets. The training set contains 14,377,989 images, while the small and large test sets contain 6,000 and 143,999 images, respectively. The small test set contains exactly 2,000 samples from each of the three languages, whereas the large test set contains 48,000 English, 47,999 Persian, and 48,000 Arabic samples. The organization and intended purposes of these partitions were described in *Section 3.6. Dataset Organization and Splits*; their language distributions are summarized here to characterize the completed dataset. Table 2 summarizes the language-specific distribution of samples across the complete dataset and each released partition.

**Table 2.** Distribution of MuSTEL images across the complete dataset and released partitions.

| Dataset | English | Persian | Arabic | Total |
|---|---|---|---|---|
| Complete dataset | 4,853,510 | 4,825,564 | 4,848,914 | 14,527,988 |
| Training dataset | 4,803,510 | 4,775,565 | 4,798,914 | 14,377,989 |
| Small test dataset | 2,000 | 2,000 | 2,000 | 6,000 |
| Large test dataset | 48,000 | 47,999 | 48,000 | 143,999 |

MuSTEL also provides substantial typographic coverage through a validated repository containing 2,009 unique font files from 711 font families. Font compatibility was assessed separately for English, Persian, and Arabic because individual font resources may support one or more writing systems. The resulting language-specific compatibility statistics for both font files and font families are summarized in Table 3.

**Table 3.** Language-specific font compatibility in the MuSTEL repository.

| Statistic | English | Persian | Arabic | Overall |
|---|---|---|---|---|
| Compatible font files | 1,957 | 1,976 | 1,927 | 2,009 unique files |
| Compatible font families | 709 | 711 | 711 | 711 unique families |

## 4.2. Textual Statistics

MuSTEL contains both monolingual and mixed-language text-line images. Monolingual samples contain Persian, Arabic, or English, whereas mixed-language samples contain two or three languages already present in the source text. Language composition was identified using the source-language label and character-level analysis of the transcription, since Persian and Arabic share many Unicode characters and cannot always be distinguished from script information alone.

Monolingual English, Arabic, and Persian samples account for the majority of the dataset, while 1,871,326 samples (12.88%) contain more than one language within the same text line. The full distribution of language-composition categories is reported in Table 4.

**Table 4.** Distribution of monolingual and mixed-language samples in the complete MuSTEL dataset.

| Language composition | Number of samples | Percentage |
|---|---|---|
| Persian | 3,357,964 | 23.11% |
| Arabic | 4,468,877 | 30.76% |
| English | 4,829,830 | 33.25% |
| Mixed-language | 1,871,317 | 12.88% |
| Total | 14,527,988 | 100.00% |

Persian-English and Arabic-English compositions combine right-to-left and left-to-right writing systems, Persian-Arabic samples combine two related but linguistically distinct Arabic-script languages, and trilingual samples contain all three target languages within a single transcription.

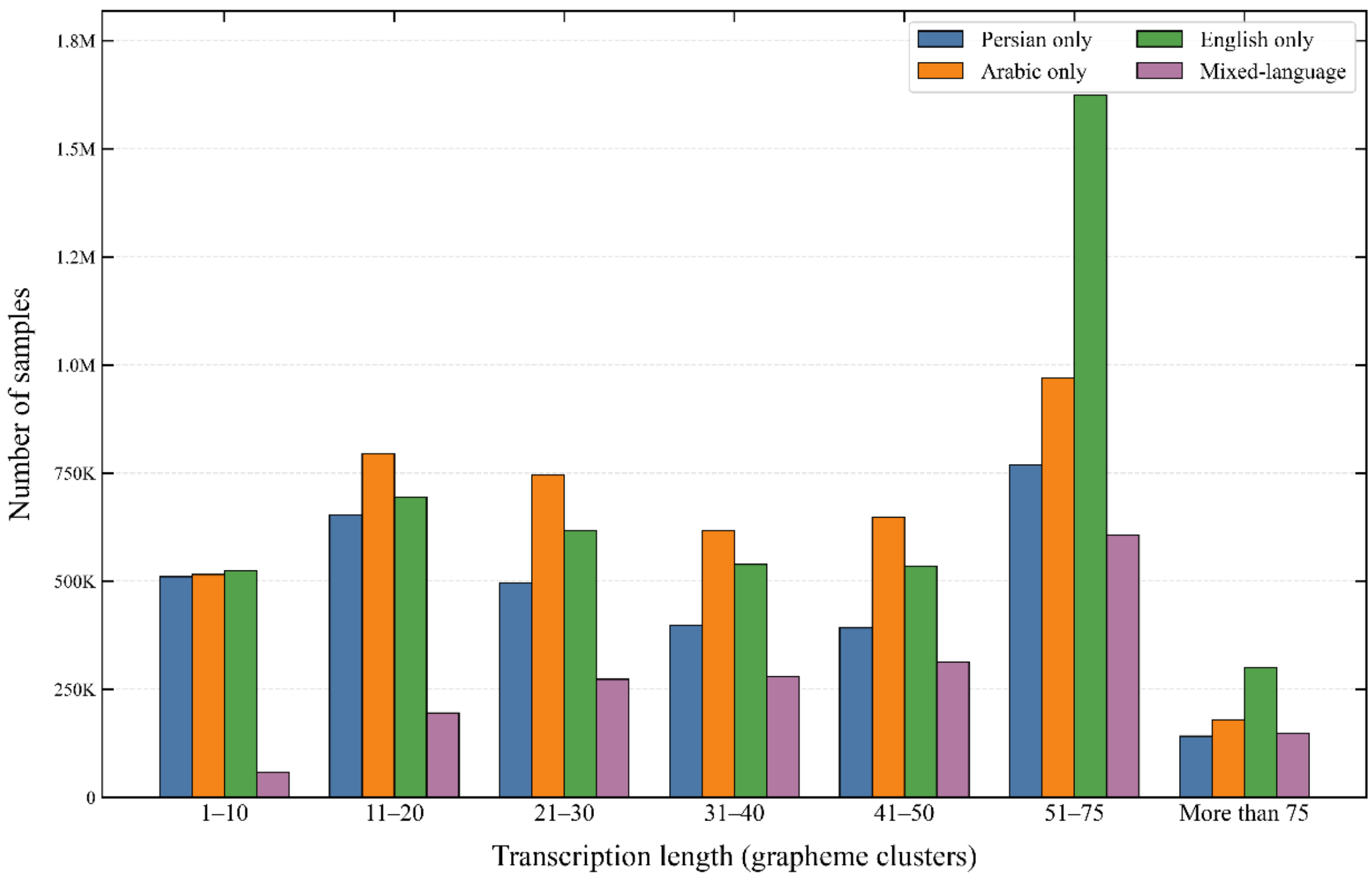

**Figure 5.** Distribution of transcription lengths in the MuSTEL dataset.

Transcription length was measured using Unicode grapheme clusters rather than simple code-point counts. A grapheme cluster treats a letter and its attached marks as one unit; for example, «بَ» is counted as one unit rather than counting «ب» and the vowel mark «َ» separately. Whitespace characters were included because spaces form part of the stored OCR transcription and contribute to the effective sequence length processed by recognition models.

The analysis shows that the mean transcription lengths for Persian, Arabic, English, and mixed-language samples are 34.67, 35.76, 41.77, and 45.08 characters, respectively, with an overall mean of 38.71. Mixed-language samples have the longest transcriptions on average, while Persian samples have the shortest. Figure 5 presents the transcription-length distributions across these groups.

Words were defined as contiguous sequences of Unicode letters, combining marks, and numerals. Punctuation was treated as a token boundary, while numerical sequences were retained as valid tokens. Unique tokens were counted in their original written forms to characterize the orthographic diversity relevant to OCR. English capitalization was therefore preserved, with forms such as “Book” and “book” counted separately. Persian and Arabic diacritics were also retained, so forms differing in the presence or type of diacritics were counted as distinct tokens.

Across the complete dataset, MuSTEL contains more than 103 million word occurrences and approximately 5.2 million unique words, indicating substantial lexical diversity. Mixed-language samples have the highest mean number of words per sample among the analyzed groups. Table 5 reports the word statistics by language composition. Figure 6 further illustrates the most frequent words observed in the Persian, Arabic, and English portions of the dataset.

**Table 5.** Word-level statistics of the MuSTEL dataset.

| Sample group | Total word occurrences | Unique words | Mean words per sample |
|---|---|---|---|
| Persian only | 24,250,427 | 831,753 | 7.22 |
| Arabic only | 29,371,668 | 2,462,559 | 6.57 |
| English only | 34,708,380 | 1,463,709 | 7.19 |
| Mixed-language | 15,584,088 | 1,420,786 | 8.33 |
| Overall | 103,914,563 | 5,196,665 | 7.15 |

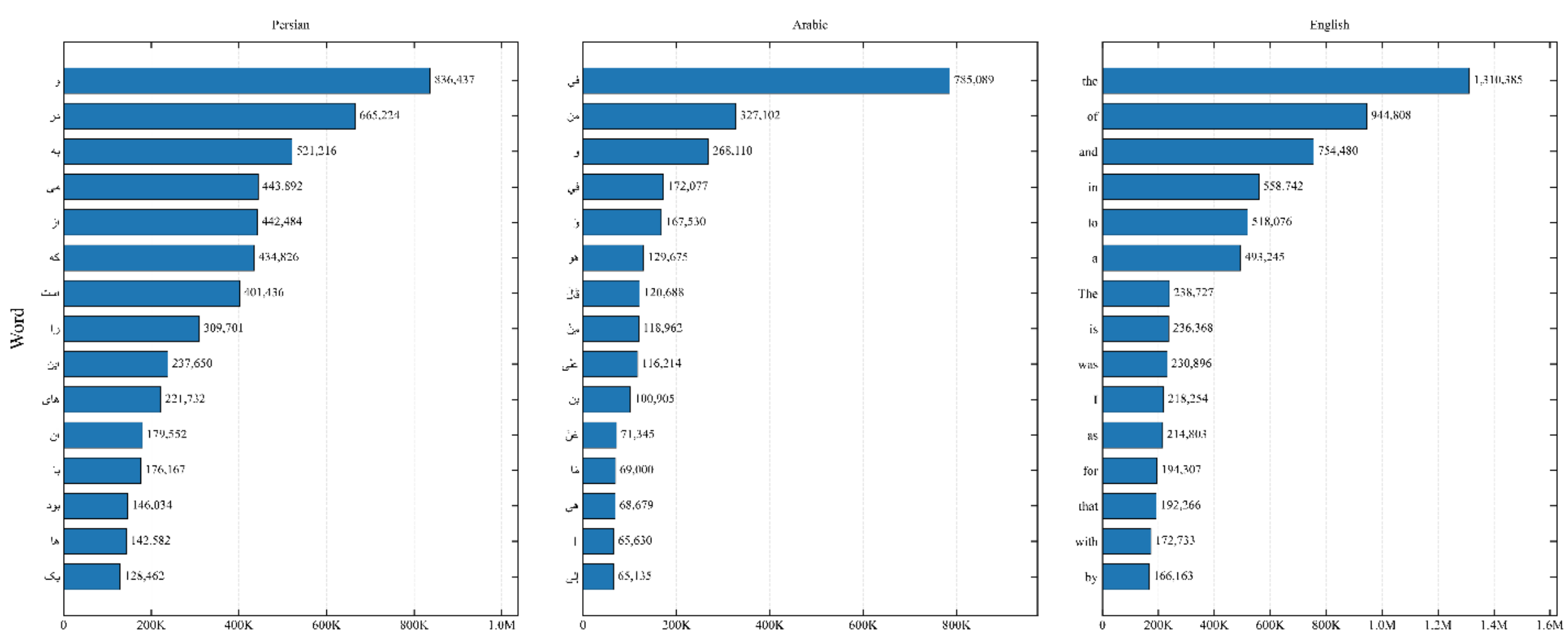

**Figure 6.** Most frequent Persian, Arabic, and English words in the MuSTEL dataset.

Character n-grams were extracted using exact Unicode code-point sequences. Each transcription was first divided into non-whitespace segments, preventing n-grams from crossing spaces or line boundaries. Punctuation was retained within non-whitespace segments and could therefore participate in the extracted sequences.

As summarized in Table 6, the number of distinct sequences increases substantially with n-gram order, reaching 675,457 unique trigrams. A large proportion of bigrams and trigrams occur only once, while many others appear more than 100 times, indicating a combination of long-tail sequence diversity and frequently recurring orthographic patterns. Figure 7 presents the most frequent character bigrams and trigrams across the dataset.

**Table 6.** Character n-gram statistics of the MuSTEL dataset.

| N-gram order | Total occurrences | Unique n-grams | Occurring once | Occurring more than 100 times |
|---|---|---|---|---|
| Unigrams | 518,972,921 | 6,821 | 1,308 | 938 |
| Bigrams | 418,557,980 | 130,447 | 69,693 | 11,220 |
| Trigrams | 323,313,177 | 675,457 | 300,059 | 69,056 |

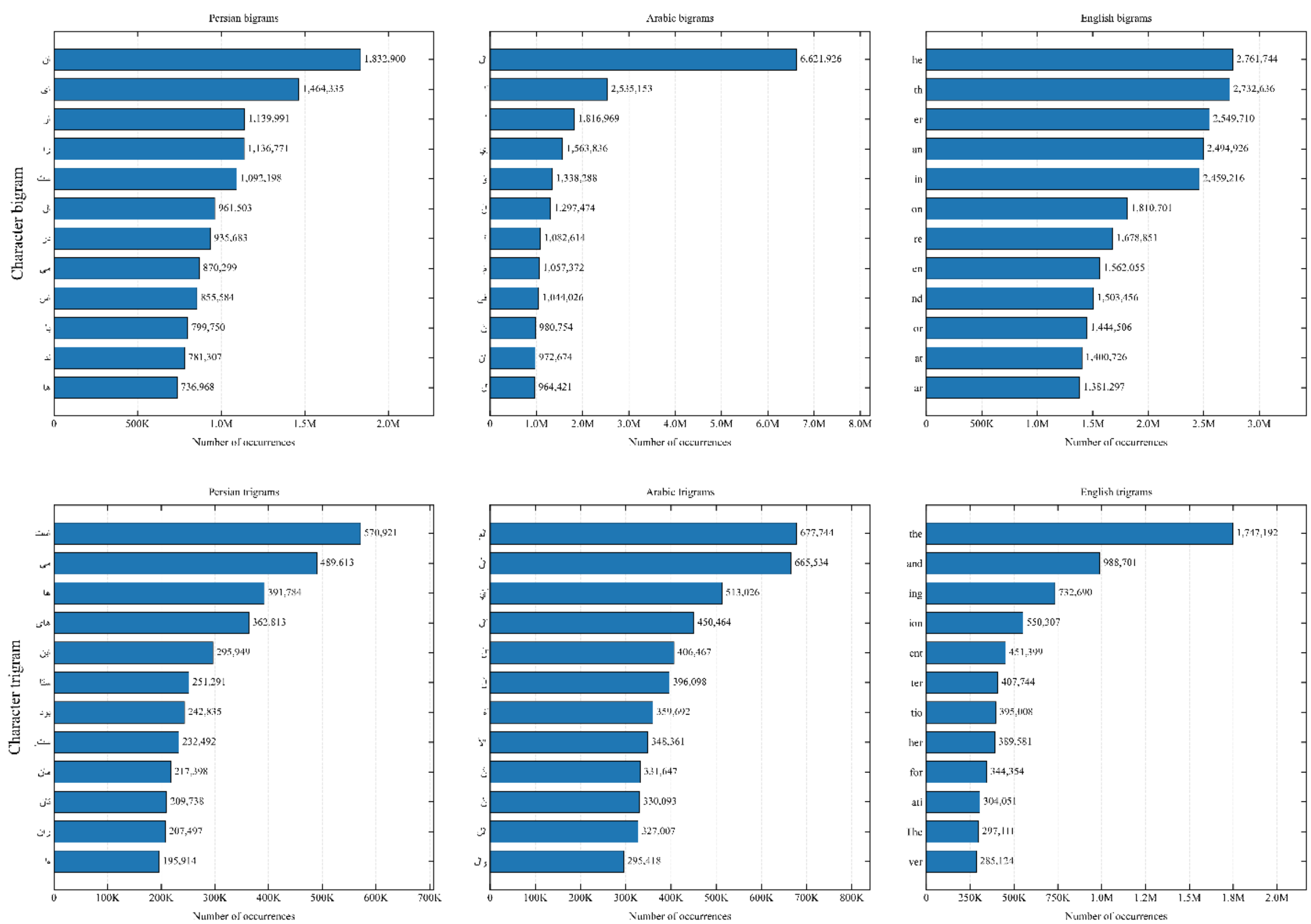

**Figure 7.** Most frequent character bigrams and trigrams in the MuSTEL dataset.

## 4.3. Image Statistics

Image width and height were obtained directly from the generated files and are reported in pixels. The aspect ratio was calculated as the image width divided by its height.

Image dimensions vary substantially across MuSTEL, particularly in width. As summarized in Table 7, horizontal variation is considerably greater than vertical variation, reflecting differences in transcription length, glyph width, and typography. Image height is comparatively more stable because it is influenced mainly by font size, font metrics, diacritics, and margins.

**Table 7.** Image-dimension statistics for the MuSTEL dataset.

| Dimension | Mean | Standard deviation |
|---|---|---|
| Width (pixels) | 455.47 | 379.96 |
| Height (pixels) | 39.85 | 14.26 |
| Aspect ratio | 11.22 | 7.46 |

The median image width is 354 pixels, with the central 50% of samples ranging from 198 to 592 pixels. Ninety percent of images have widths between 78 and 1,182 pixels. Image height is more concentrated, with a median of 36 pixels, a central 50% range of 31–42 pixels, and a 90% range of 24–74 pixels. The median aspect ratio is 9.64, with a central 50% range of 5.60–15.17. These distributions are right-skewed, particularly for width and aspect ratio, indicating the presence of a relatively small number of very wide text-line images. Figure 8 illustrates the distributions of image width and height across the complete dataset.

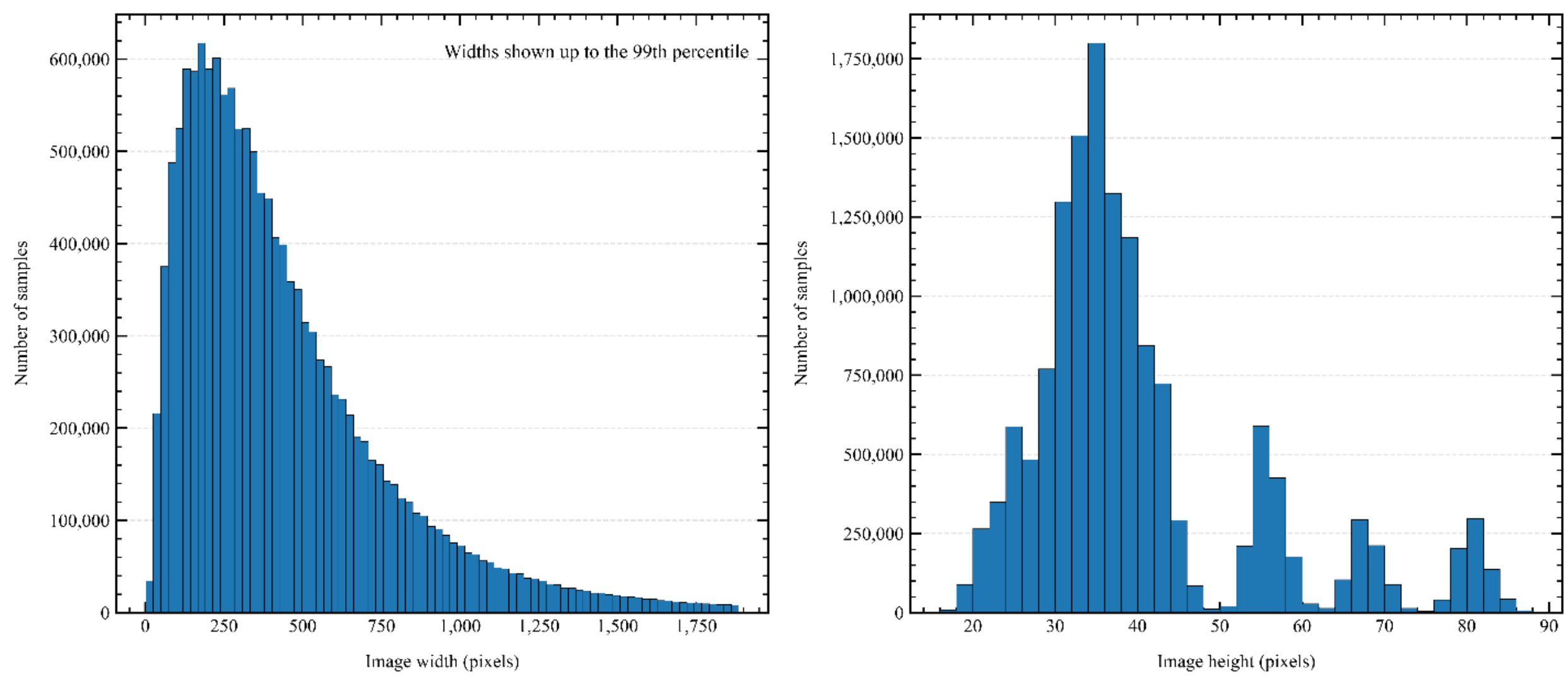

**Figure 8.** Distributions of image width and height in the MuSTEL dataset.

Transcription length and image width are positively related. Across all samples, the Pearson correlation coefficient is 0.652, indicating a moderate-to-strong relationship. Longer transcriptions generally produce wider images, although width is also affected by script, font family, font size, spacing, margins, style, and multiple-font composition. This relationship is visualized in Figure 9.

Font and style diversity was analyzed using the font-family, font-file, and style assignments stored in the metadata of all samples. Every font file and family in the validated repository appears at least once in the dataset.

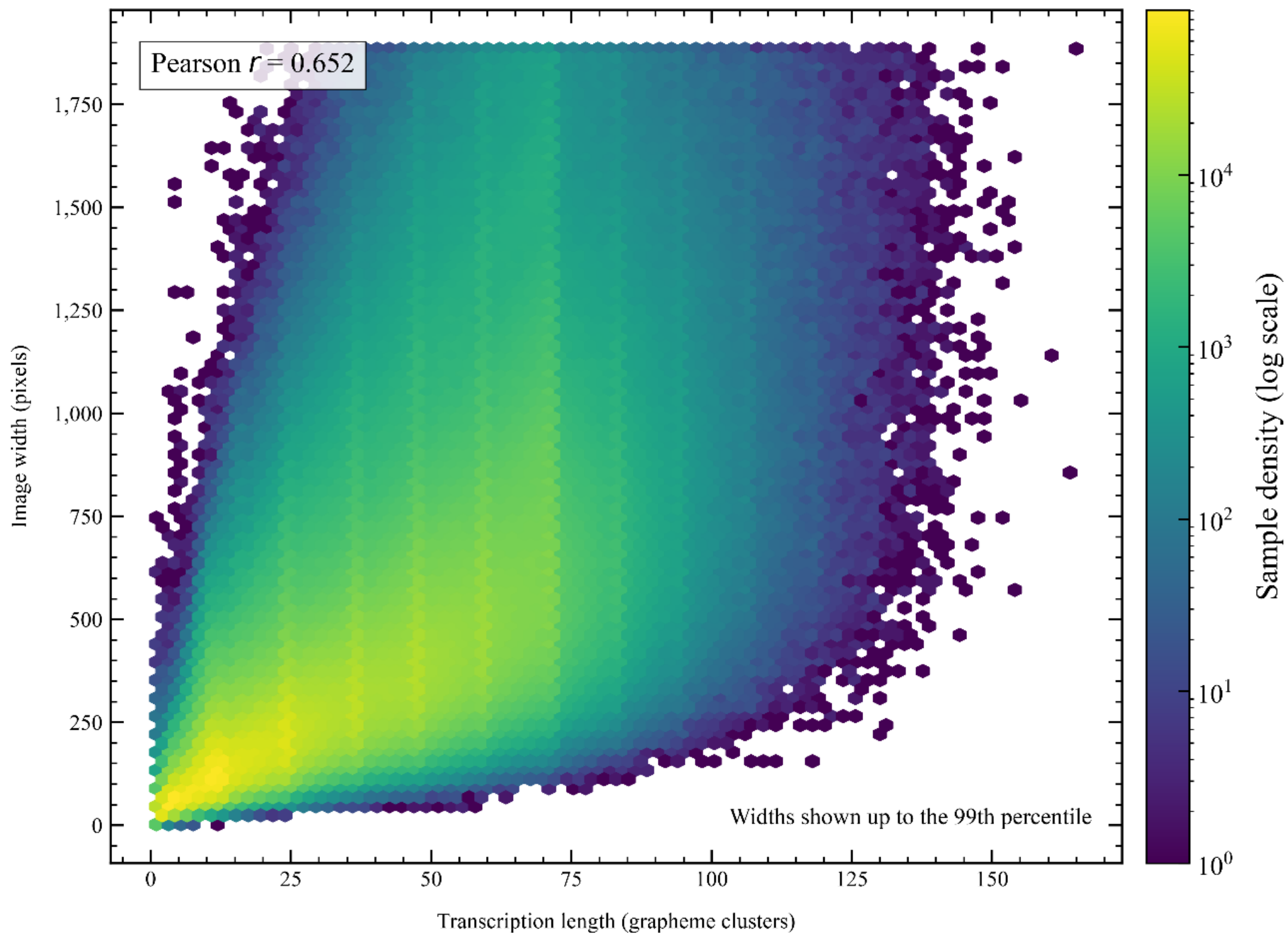

**Figure 9.** Relationship between transcription length and image width in the MuSTEL dataset.

Font-family usage was measured from the recorded rendering assignments. Because lines containing multiple font runs contribute more than one assignment, the total number of font-family assignments exceeds the number of images. Usage is distributed across a broad range of families rather than being concentrated in only a few dominant typefaces. Figure 10 shows the most frequently used font families, with Arial having the largest individual share but accounting for less than 3% of all assignments.

Style diversity was analyzed from the recorded "font_style" labels, which were consolidated into seven broader categories for analysis. Regular and Bold account for most style assignments, while Light, Italic/Oblique, Medium, Bold Italic/Oblique, Condensed, and other categories provide additional stylistic variation. The complete grouped statistics are reported in Table 8.

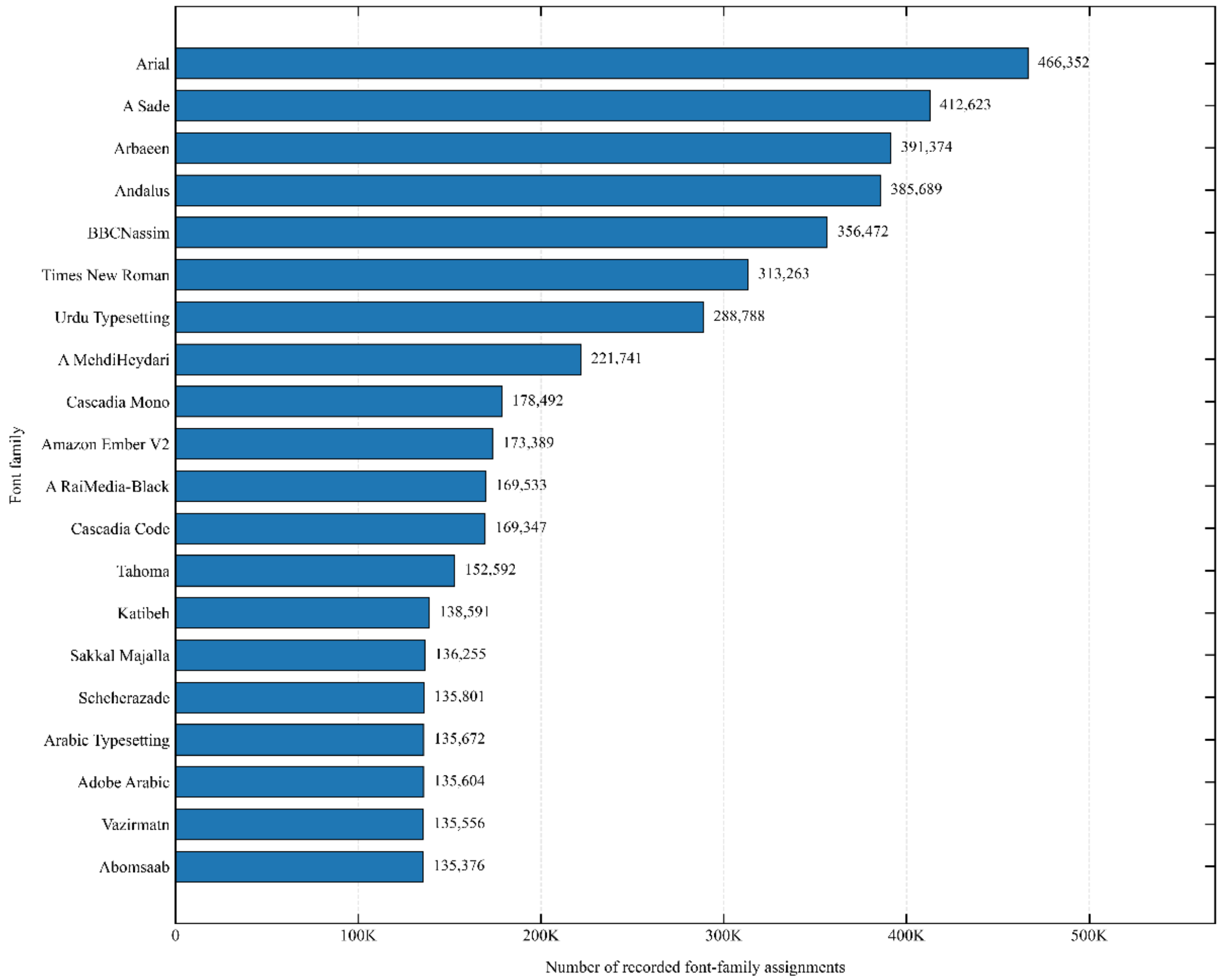

**Figure 10.** Most frequently used font families in the MuSTEL dataset.

**Table 8.** Grouped distribution of font styles in the MuSTEL dataset.

| Style category | Observed font files supporting style | Recorded assignments | Percentage |
|---|---|---|---|
| Regular | 567 | 10,936,247 | 67.57% |
| Bold | 503 | 3,177,363 | 19.63% |
| Italic/Oblique | 294 | 468,086 | 2.89% |
| Bold Italic/Oblique | 223 | 177,774 | 1.10% |
| Light | 272 | 834,986 | 5.16% |
| Medium | 126 | 398,367 | 2.46% |
| Other recorded categories | 24 | 192,173 | 1.19% |
| Total | — | 16,184,996 | 100.00% |

MuSTEL contains both typographically uniform and varied text lines. Most samples use one font family and style, while approximately 8% include two families or two styles. Table 9 reports the detailed distribution.

Four mutually exclusive background categories occur in the completed dataset: Gaussian noise, plain white, quasicrystal, and image backgrounds selected from a folder of background images. Their frequencies are uneven, with plain white backgrounds accounting for the largest share and quasicrystal backgrounds the smallest. The distribution is also similar across Persian, Arabic, and English samples, indicating consistent background selection across the three primary languages. Figure 11 shows the overall distribution of these four categories.

**Table 9.** Distribution of MuSTEL samples by the number of distinct font families and styles per line.

| Distinct types per line | Font Families | | Styles | |
|---|---|---|---|---|
| | Number of samples | Percentage | Number of samples | Percentage |
| One | 13,393,197 | 92.19% | 13,370,065 | 92.03% |
| Two | 1,134,791 | 7.81% | 1,157,923 | 7.97% |

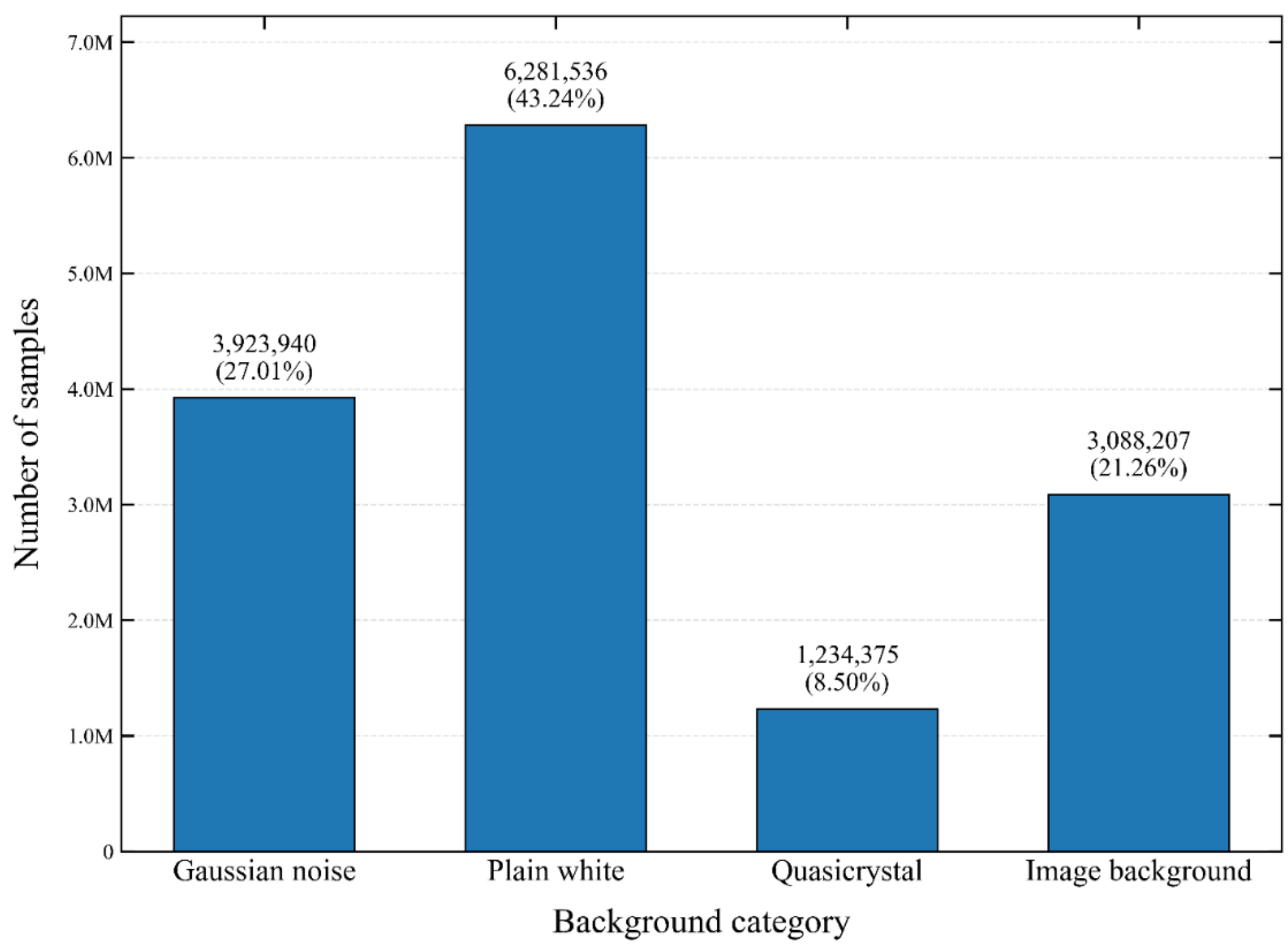

**Figure 11.** Distribution of background categories in the MuSTEL dataset.

Rendering variation was considered active when the corresponding metadata field differed from its default value. As summarized in Table 10, reduced image quality and blur are the most prevalent conditions, while geometric distortion is present in approximately half of the dataset. Stroke and skew occur less frequently, and highlighting and underlining each appear in roughly 7% of samples. Because multiple rendering conditions may occur in the same image, the percentages in Table 10 are not mutually exclusive and should not be summed.

The rendering settings are drawn from controlled discrete parameter ranges, including several blur, skew, stroke-width, and image-quality levels. This produces samples with varying degrees of visual modification while preserving the corresponding ground-truth transcription. Overall, the statistics indicate that MuSTEL contains substantial image-appearance variation rather than being dominated by uniformly rendered text lines.

**Table 10.** Prevalence of rendering variations in the MuSTEL dataset.

| Rendering variation | Number of samples | Percentage of all samples |
|---|---|---|
| Reduced image save quality* | 12,264,797 | 84.42% |
| Blur | 11,213,088 | 77.18% |
| Geometric distortion | 7,265,334 | 50.01% |
| Stroke | 3,704,640 | 25.50% |
| Skew | 1,613,904 | 11.11% |
| Highlight | 1,004,701 | 6.92% |
| Underline | 1,003,863 | 6.91% |
| Character spacing | 19 | <0.01% |

* Reduced image save quality refers to samples saved using a lower quality setting, independently of blur, distortion, and other rendering variations.

## 5. Discussion

The construction and statistical analysis of MuSTEL indicate that the dataset addresses several limitations identified for existing OCR resources, particularly in multilingual line-level recognition involving Persian, Arabic, and English. Its main strengths arise from the combination of large scale, multilingual composition, typographic diversity, script-aware generation, and controlled rendering variation. At the same time, the synthetic nature and line-level scope of the dataset introduce limitations that should be considered when it is used in future OCR studies.

### 5.1. Advantages of MuSTEL

MuSTEL combines large scale with multilingual coverage, providing 14,527,988 text-line images with broadly balanced representation of Persian, Arabic, and English, including 1,871,326 mixed-language samples. As summarized in Table 1, MJSynth and SynthText primarily provide English synthetic data, while APTI and SARD focus on Arabic, and Shotor and IDPL-PFOD on Persian. MuSTEL extends this coverage by supporting all three languages, including lines whose source text contains multiple languages and writing directions.

Another advantage is its typographic diversity. MuSTEL uses 2,009 validated font files from 711 families, all represented in the completed dataset. Compared with the smaller font collections used by resources such as APTI and IDPL-PFOD, this repository provides broader typographic coverage, together with within-line font-family and style changes. Capability-aware selection assigns fonts according to their validated support for the required letters, numerals, punctuation, hamzah forms, and diacritics.

MuSTEL also provides substantial lexical and visual variability, with more than 103 million word occurrences, over 5.19 million unique words, and extensive character bigram and trigram inventories. Its rendering conditions include varied backgrounds, blur, geometric distortion, skew, stroke width, image save quality, highlighting, and underlining. Although existing synthetic resources also incorporate typographic or visual variation, MuSTEL combines these properties with multilingual text-line content in a single dataset. Its line-level focus complements scene-text and page-level document resources.

### 5.2. Potential Applications

MuSTEL is intended primarily for training and evaluating line-level OCR systems for Persian, Arabic, English, and multilingual documents. It can support recognition pipelines in which document pages are first segmented or detected into text lines and the extracted lines are subsequently transcribed. This setting is relevant to books, reports, archives, digitized collections, and other text-dense documents.

The dataset may also support research on mixed-direction recognition, script-aware OCR, multilingual sequence modeling, robustness to font variation, and recognition under different rendering conditions. Because generation metadata includes language composition, typography, image dimensions, backgrounds, and rendering parameters, researchers can additionally construct controlled subsets for analyzing specific OCR challenges.

### 5.3. Limitations and Future Work

The principal limitation of MuSTEL is that it is synthetic. Although the generation process introduces diverse typography, backgrounds, geometric variation, blur, and image-quality conditions, synthetic rendering cannot reproduce every characteristic of real scanned or photographed documents. Real documents may contain irregular illumination, physical damage, paper texture, handwriting, complex scanning artifacts, or layout interactions that are not fully represented.

MuSTEL is also restricted to line-level images and therefore does not model complete page layouts, text detection, reading-order prediction across regions, tables, figures, or other document-structure problems. In addition, the dataset currently focuses on Persian, Arabic, and English, so its multilingual coverage does not extend to other writing systems.

Further development could expand the language inventory, increase the diversity of document-like visual conditions, and introduce additional real or synthetic document sources. Future versions may also extend beyond isolated text lines toward page-level resources while retaining the script-aware font validation, multilingual composition, and structured metadata framework developed for MuSTEL.

## 6. Conclusion

This paper introduced MuSTEL, a large-scale synthetic multilingual dataset for line-level OCR of Persian, Arabic, English, and mixed-language text. The dataset was developed to address limitations of existing OCR resources related to multilingual coverage, typographic diversity, within-line language and font variation, and document-like visual conditions.

MuSTEL contains 14,527,988 annotated text-line images generated from multilingual natural-language corpora and organized into a large training set and two predefined evaluation sets. Its generation framework extends TRDG with script-aware text processing, bidirectional composition, capability-based font selection, multiple-font and style rendering, and automated annotation and quality-control procedures. The validated font repository contains 2,009 font files from 711 families, and the completed dataset includes substantial lexical and character-sequence diversity together with varied image dimensions, backgrounds, and rendering conditions.

The statistical analysis shows that MuSTEL provides broad coverage across the three primary languages while also including a substantial number of mixed-language samples. Its combination of linguistic, typographic, and visual diversity makes it suitable for research on multilingual and mixed-direction line recognition, particularly in document-oriented OCR pipelines. At the same time, its synthetic and line-level nature means that it does not fully reproduce all properties of real scanned or photographed documents or complete page layouts. Future work should therefore examine transfer to real-world OCR data, evaluate modern recognition architectures using the released partitions, and extend the framework toward additional languages, visual conditions, and page-level resources.